\documentclass[screen,acmconf]{acmartedit}
\usepackage[utf8]{inputenc}
\usepackage{amssymb,amsmath,amsfonts,amsthm}
\usepackage{color}
\usepackage{graphicx}
\usepackage{tikz}
\usetikzlibrary{decorations.pathreplacing}
\usepackage{sidecap}
\usepackage{subfig}
\usepackage{algorithm,algorithmic}
\usepackage{setspace}

\usepackage{xpatch}

\renewcommand\footnotetextcopyrightpermission[1]{} 
\setcopyright{none}
\renewcommand\footnotetextauthorsaddresses[1]{} 
\title{Fooling Computer Vision into Inferring the Wrong Body Mass Index}

\author{Owen Levin, Zihang Meng, Vikas Singh and Xiaojin Zhu}
\affiliation{%
  \institution{University of Wisconsin--Madison}
}

\date{Date}
\begin{document}
\maketitle
\begin{abstract}
Recently it's been shown that neural networks can use images of human faces to accurately predict Body Mass Index (BMI), a widely used health indicator. In this paper we demonstrate that a neural network performing BMI inference is indeed vulnerable to test-time adversarial attacks. This extends test-time adversarial attacks from classification tasks to regression.
The application we highlight is BMI inference in the insurance industry, where such adversarial attacks imply a danger of insurance fraud.
\end{abstract}
\section{Introduction}
Body Mass Index (BMI) is a widely used health quantity calculated as $kg / m^2$.
The world health organization categorizes BMI broadly into
Underweight $[0,18.5)$, Normal $[18.5,25)$, Overweight $[25,30)$, and Obese $[30,+\infty)$~\cite{BMItable}.
Kocabey et al. recently developed a regression task Face-to-BMI~\cite{Face2BMI}, where they accurately predicted BMI from images of human faces. The motivation for their study was identifying how an individual's BMI affects their treatment by others on social media platforms~\cite{BMIsocialMedia}. 

In this paper we instead focus on the application of Face-to-BMI in the insurance industry, where adversarial attacks could become a issue. 
Suppose an insurance company uses a neural network to predict the BMI of their clients from photos and then uses this information to influence coverage.
There are two scenarios in which an adversarial attacker may want to manipulate the input photo inperceptibly to attack the BMI predictor: 
(1) the attacker may want to make someone appear healthier to lower their rates;  
(2) conversely, make someone appear unhealthy to sabotage that person's insurance application.
We demonstrate that a neural network performing Face-to-BMI is indeed vulnerable to test-time adversarial attacks.  
This extends test-time adversarial attacks from classification tasks (e.g.~\cite{l0example,l2example,linfexample,attackexample}) to regression.
%
%
%
\section{Adversarial Attacks on Face-to-BMI Prediction}\label{Attack}

\textbf{The victim neural network}  
$f:\mathbb{R}^{227\times 227\times3}\to\mathbb{R}$ 
takes as input a $227\times227\times3$ face image and outputs a BMI estimate.  
We use Alexnet~\cite{krizhevsky2012imagenet} layers conv1 to fc7 plus one linear layer after fc7 to perform regression.

\textbf{The threat model} assumes a whitebox attacker with full knowledge of the victim weights and architecture.
The attacker can edit any pixels in the photo, including those not on the human.  
We consider targeted attacks to force $f$ prediction into a pre-specified target range $[L,U]\subset\mathbb{R}$.

\textbf{The attack formulation}
find the minimum perturbation $\delta$ such that for input input $X$, $f(X+\delta)\in[L,U]$.  
Both $X$ and $X+\delta$ must be valid images with integer pixel values in 0--255.
We measure perturbation by its $\ell_p$ norm $\|\delta\|_p$ for some $p\in(0,\infty]$~\cite{l0example,l2example,linfexample,attackexample}. 
Thus, the ideal attack solves

{\fontsize{8}{0}\selectfont 
\begin{equation}\label{eq:trueProblem}
    \displaystyle\min_{\delta\in\mathbb{R}^{227\times 227\times 3}}
        \ \   \|\delta\|_p
   \ \ \text{subject to}
        \ \   L  \leq f(X+\delta)  \leq    U,\ \ \text{and}
        \ \ 
           (X+\delta)\in I:=\{0,\ldots,255\}^{227\times 227\times3}.
\end{equation}
}
However, this is a difficult integer program.
We heuristically solve a related problem to simply find a \emph{small enough} $\delta$. 
We reformulate the attack goal as follows:
$     L \leq f(X+\delta)   \leq U
    \Leftrightarrow 
    \left(f(X+\delta)-\frac{U+L}{2}\right)^2    
        \leq \left(\frac{U-L}{2}\right)^2.
$
We relax the integral constraint on $\delta$ and change the objective:
{\fontsize{8}{0}\selectfont 
\begin{equation}\label{eq:relaxedProblem}
    \displaystyle\min_{\delta\in {\mathbb R}^{227\times 227\times3}}
        \ \    \left(f(X+\delta)-\frac{U+L}{2}\right)^2
   \ \ \text{subject to}
        \ \   (X+\delta)\in [0,255]^{227\times 227\times3}.
\end{equation}
}
We initialize $\delta=\mathbf{0}$ and perform early-stopping as soon as $f(X+\mathrm{Round}(\delta)) \in [L, U]$ to encourage small norm on $\delta$.

\section{Experiments}

\textbf{Datasets}.
We use two datasets of (photo, BMI) pairs: (1) Federal Corrections Body Mass Index (FCBMI) consists of 9045 public photos at multiple federal and state corrections facilities. 
(2) VisualBMI dataset with 4206 photos collected by~\cite{Face2BMI} from Reddit.

\textbf{Training the victim network}.
We train the BMI prediction network with transfer-learning. 
We load weights pre-trained on the ILSVRC 2012 data set for the conv1 to fc7 layers of Alexnet. Then we randomly initialize the last linear layer using Xavier~\cite{glorot2010understanding}.  Finally we fine tune the entire network's weights using our own training images.
We use a random subset of 7000 images in FCBMI for fine-tuning, and keep the remaining 2045 images in FCBMI and the whole VisualBMI for testing.
We pre-process the images identically to in AlexNet~\cite{krizhevsky2012imagenet}: images are converted from RGB to BGR, re-sized to $227\times227\times 3$. Finally we subtract the grand mean pixel value from each pixel in the images in the training set. This means that we provide an input in $[-255,255]^{227\times 227\times3}$ to the neural network at test time.
During training we use $\ell_2$ loss. We use the Adam~\cite{kingma2014adam} optimizer with $\beta_1=0.9, \beta_2=0.999$. The batch size is 64 and learning rate is 0.0001.

\textbf{Attack implementation}.
To solve~\eqref{eq:relaxedProblem} the attacker simulates the victim by pre-pending an extra input layer with $X$ and 1s:
    \begin{center}\begin{tikzpicture}[xscale=1,yscale = .5]
        \foreach \x in {1,...,2}
        {
            \pgfmathtruncatemacro{\nodelabel}{\x}
            \node[circle,draw=black,fill=white!80!black,minimum size=15]
                (x\nodelabel)
                at (0,1-\x) {{\scriptsize $X_\nodelabel$}};
        }
        \node[minimum size=15]
                at (0,1-3) {{\tiny $\vdots$}};
        \node[circle,draw=black,fill=white!80!black,minimum size=15]
                (xn)
                at (0,1-4) {{\scriptsize $X_n$}};
        \foreach \d in {1,2}
        {
        \pgfmathtruncatemacro{\nodelabel}
        {\d}
            \node[circle,draw=black,fill=white!80!black,minimum size=15]
                (d\nodelabel)
                at (0,1-4-\d) {$1$};
        }
        \node[minimum size=15]
                at (0,1-4-3) {{\tiny $\vdots$}};
        \node[circle,draw=black,fill=white!80!black,minimum size=15]
                (dn)
                at (0,1-4-4) {$1$};
        \foreach \xd in {1,2}
        {
        \pgfmathtruncatemacro{\nodelabel}
        {\xd}
            \node[circle,draw=black,fill=white!80!black,minimum size=15]
                (p\nodelabel)
                at (3,1-1.75*\xd) {{\scriptsize $\sum$}};
            \node[circle,draw=black,fill=white!80!black,minimum size=15]
                (xd\nodelabel)
                at (4.25,1-1.75*\xd) {};
        }
        \node[minimum size=15]
                at (3,1-1.75*3) {{\scriptsize $\vdots$}};
        \node[minimum size=15]
                at (4.25,1-1.75*3) {{\scriptsize $\vdots$}};
        \node[circle,draw=black,fill=white!80!black,minimum size=15]
                (pn)
                at (3,1-1.75*4) {{\scriptsize $\sum$}};
        \node[circle,draw=black,fill=white!80!black,minimum size=15]
                (xdn)
                at (4.25,1-1.75*4) {};
        \foreach \i in {1,2,n}
        {
            \draw (p\i) -- node[below]{\footnotesize $1$} (xd\i);
            \draw (x\i) -- node[below] {\footnotesize $1$} (p\i);
            \draw (d\i) -- node[below] {\footnotesize $\delta_\i$} (p\i) ;
        }
        \draw[draw=black,fill=white!80!black] (5.5,-2) rectangle (9.5,-4);
        \node at (7.5,-2.5) {\scriptsize Preprocessing +};
        \node at (7.5,-3) {\scriptsize AlexNet+Linear regression};
        \node at (7.5,-3.5) {(frozen weights)};
        \foreach \i in {1,2,n}
        {
            \draw (xd\i) -- (5.5,-3);
        }
        \draw (9.5,-3) --(10.75,-3);
        \node[circle,draw=black,fill=white!80!black,minimum size=15,text width=30,align=center] at (10.75,-3) {\scriptsize Predicted BMI};
        \draw[decorate,decoration={brace,amplitude=10pt,mirror}]
(-.5,.5) -- (-.5,-3.5) node [black,midway,xshift=-0.6cm] 
{$X$};        
        \draw[decorate,decoration={brace,amplitude=10pt,mirror}]
        (-.5,-3.5) -- (-.5,-7.5) node [black,midway,xshift=-0.6cm] 
        {$\delta$};  
        \draw[decorate,decoration={brace,amplitude=4pt,mirror}]
        (3.75,-6.7) -- (4.75,-6.7) node [black,midway,yshift=-0.4cm] 
        {{\footnotesize $(X+\delta)$}};  
    \end{tikzpicture}\end{center}
The attacker freezes the weights of the entire network except $\delta$ and trains the network using projected gradient descent on the objective in~(\ref{eq:relaxedProblem}). 
Once training is complete, the attacker takes a final projection step and rounds $\delta$ so that $(X+\delta)\in I$.

\begin{algorithm}
    \caption{Adversarially attacking the BMI prediction network}
    \label{alg:attack}
    {\fontsize{8}{0}\selectfont 
    \begin{algorithmic}
    \REQUIRE \hspace{1.1em}$f$: BMI prediction network, \\ \hspace{3.375em}$X$: victim image, \\ \hspace{3.375em}$K>0$: Max iterations
    \ENSURE $\delta$: perturbation such that $f(X+\delta)\in[L,U]$ and $(X+\delta)\in I$
      \STATE $\delta\gets\mathbf{0}$
      \STATE $k\gets 0$
      \WHILE{$k<K$ or $f(X+\mathrm{Round}(\delta))\notin [L,U]$}
        \STATE $\delta\gets \delta-\eta_k\nabla_\delta\left(f(X+\delta)-\frac{U+L}{2}\right)^2$
        \hfill\COMMENT{gradient descent with step size $\eta_k$}
        \STATE Project $\delta$ such that $(X+\delta)\in[0,255]^{227\times 227\times 3}$
        \STATE $k\gets k+1$
      \ENDWHILE
      \STATE $\delta\gets\mathrm{Round}(\delta)$ \hfill\COMMENT{rounds $\delta$ such that $X+\delta$ is moved to the nearest point in $I$}
      \RETURN $\delta$\hfill\COMMENT{flags a failure if final $\delta$ is unsuccessful after $K$ iterations}
    \end{algorithmic}
    }
\end{algorithm}

\begin{figure}[htbp]
    \subfloat[\tiny $f(X):\ $ $17.38$ \hfill $19.39$ \hfill $21.46$ \hfill $23.49$\hfill $25.48$ \hfill $27.89$ \hfill $29.41$ \hfill $31.51\ \ $]
    {  
        \includegraphics[scale=.40,trim={35 25 261 10},clip]{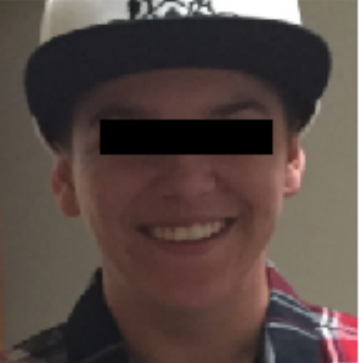}\hspace{.5em}
        \includegraphics[scale=.40,trim={35 25 261 10},clip]{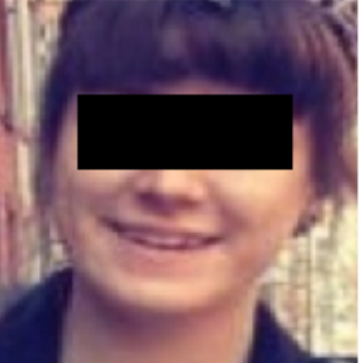}\hspace{.5em}
        \includegraphics[scale=.40,trim={35 25 261 10},clip]{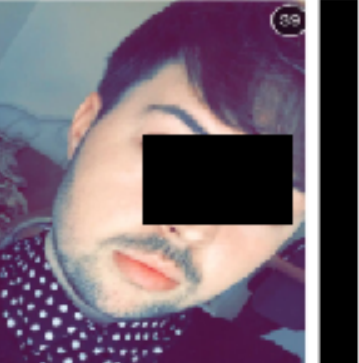}\hspace{.5em}
        \includegraphics[scale=.40,trim={35 25 261 10},clip]{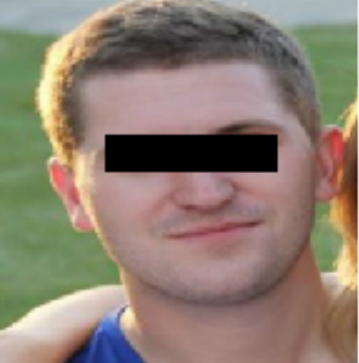}\hspace{.5em}
        \includegraphics[scale=.40,trim={35 25 261 10},clip]{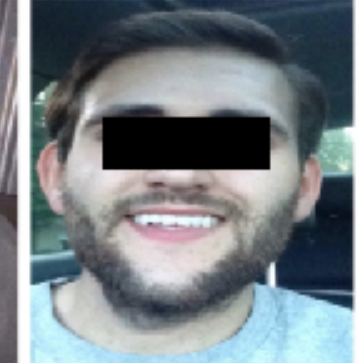}\hspace{.5em}
        \includegraphics[scale=.40,trim={35 25 261 10},clip]{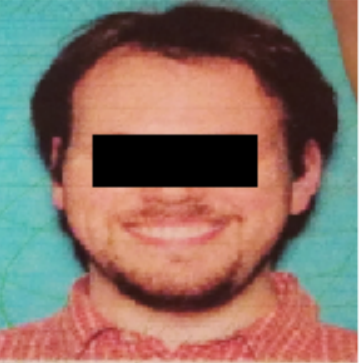}\hspace{.5em}
        \includegraphics[scale=.40,trim={35 25 261 10},clip]{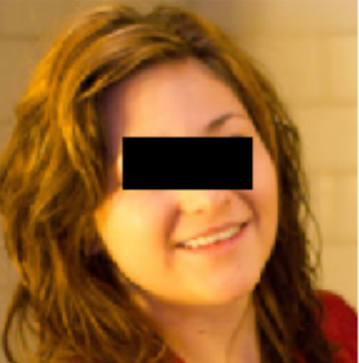}\hspace{.5em}
        \includegraphics[scale=.40,trim={35 25 261 10},clip]{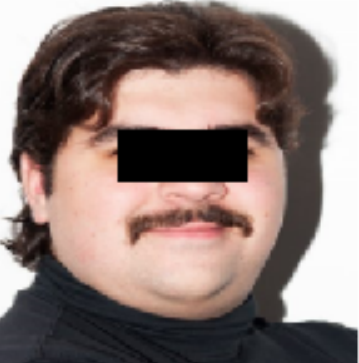}
    }\\ [1pt]
    \subfloat[\tiny $(\|\delta\|_\infty=1,\|\delta\|_2=100)$ \hfill $(0,0)$ \hfill $(0,0)$ \hfill $(0,0)$\hfill$(1,72.42)$ \hfill $(1,158.4)$ \hfill $(1,206.9)$ \hfill $(1,312.0)$]
    {  
        \includegraphics[scale=.40,trim={159 25 135 10},clip]{h17.pdf}\hspace{.5em}
        \includegraphics[scale=.40,trim={159 25 135 10},clip]{h19.pdf}\hspace{.5em}
        \includegraphics[scale=.40,trim={159 25 135 10},clip]{h21.pdf}\hspace{.5em}
        \includegraphics[scale=.40,trim={159 25 135 10},clip]{h23.pdf}\hspace{.5em}
        \includegraphics[scale=.40,trim={159 25 135 10},clip]{h25.pdf}\hspace{.5em}
        \includegraphics[scale=.40,trim={159 25 135 10},clip]{h27.pdf}\hspace{.5em}
        \includegraphics[scale=.40,trim={159 25 135 10},clip]{h29.pdf}\hspace{.5em}
        \includegraphics[scale=.40,trim={159 25 135 10},clip]{h31.pdf}
    }\\ [1pt]
    \subfloat[\tiny $f(X+\delta):\ $ $19.72$ \hfill$19.39$ \hfill$21.46$ \hfill $23.49$ \hfill $24.01$ \hfill$23.64$ \hfill$23.18$ \hfill$18.76$]
    {  
        \includegraphics[scale=.40,trim={285 25 10 10},clip]{h17.pdf}\hspace{.5em}
        \includegraphics[scale=.40,trim={285 25 10 10},clip]{h19.pdf}\hspace{.5em}
        \includegraphics[scale=.40,trim={285 25 10 10},clip]{h21.pdf}\hspace{.5em}
        \includegraphics[scale=.40,trim={285 25 10 10},clip]{h23.pdf}\hspace{.5em}
        \includegraphics[scale=.40,trim={285 25 10 10},clip]{h25.pdf}\hspace{.5em}
        \includegraphics[scale=.40,trim={285 25 10 10},clip]{h27.pdf}\hspace{.5em}
        \includegraphics[scale=.40,trim={285 25 10 10},clip]{h29.pdf}\hspace{.5em}
        \includegraphics[scale=.40,trim={285 25 10 10},clip]{h31.pdf}
    }
    \caption{\scriptsize Attacks forcing BMI predictions into the ``normal weight'' range [18.7, 24.9].
    Row (a): Original BMI prediction $f(X)$. 
Row (b): Attack $\delta$ and its norms.
    $\delta$'s color scale maps [-2, 2] linearly to [0, 255] (gray = no attack). 
Row (c): Attacked BMI prediction $f(X+\delta)$.}
\label{fig:h}
\end{figure}

\textbf{Qualitative results.}
Figure~\ref{fig:h} shows the BMI attack on 8 photos from the VisualBMI data set.
We obscured the eyes with black boxes to preserve partial anonymity of those pictured. The boxes are not present in the original data set, so neither the prediction network nor the attacker saw or were influenced by them. 
Here the attack goal is to force BMI predictions into the normal range $[L,U]=[18.7,24.9]$.
The attacker succeeds at this.
We note that all changes have small infinite norm: $\|\delta\|_\infty\leq2$. 
Also, $\delta$s have more nonzero elements and vary more the further the original BMI is from the target range.

\begin{figure}[htbp]
    \includegraphics[width=.49\linewidth,height=.3\linewidth,trim={5 5 0 0},clip]{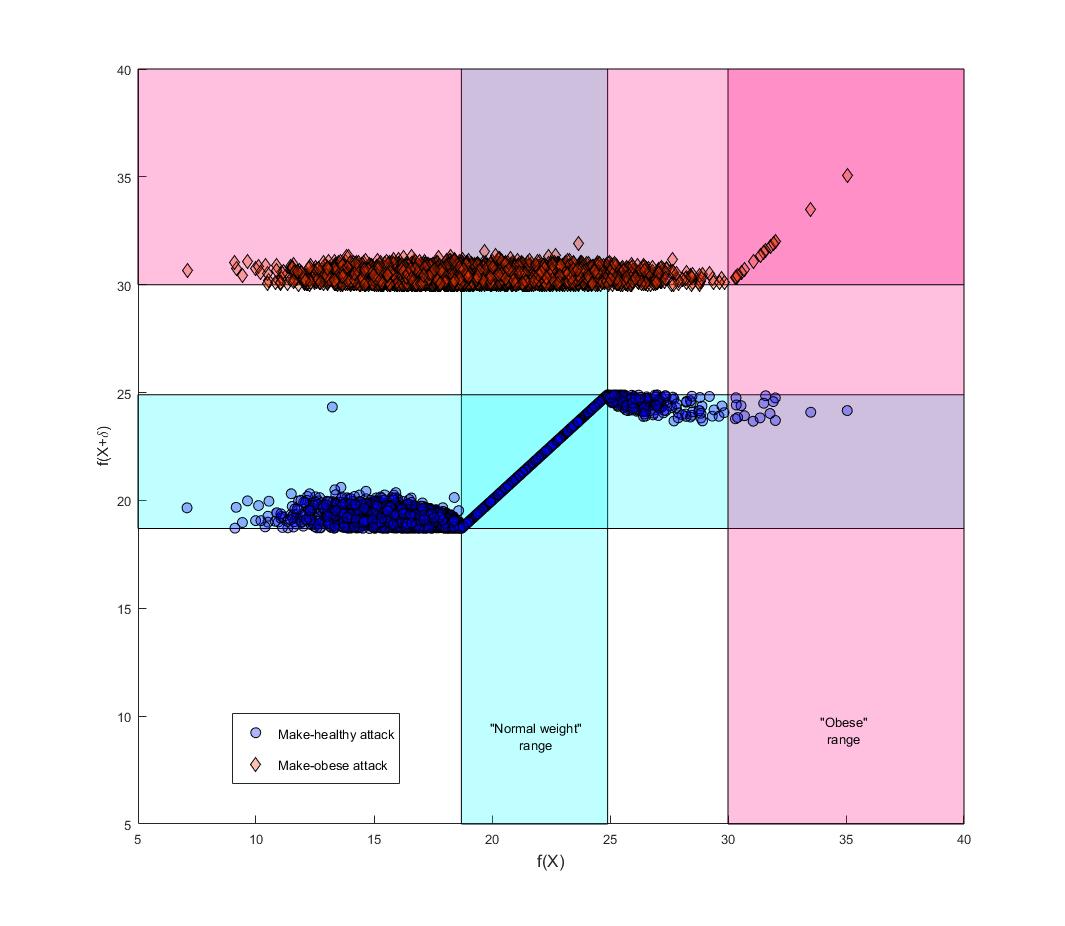}
    \includegraphics[width=.49\linewidth, height=.3\linewidth,trim={0 10 10 20},clip]{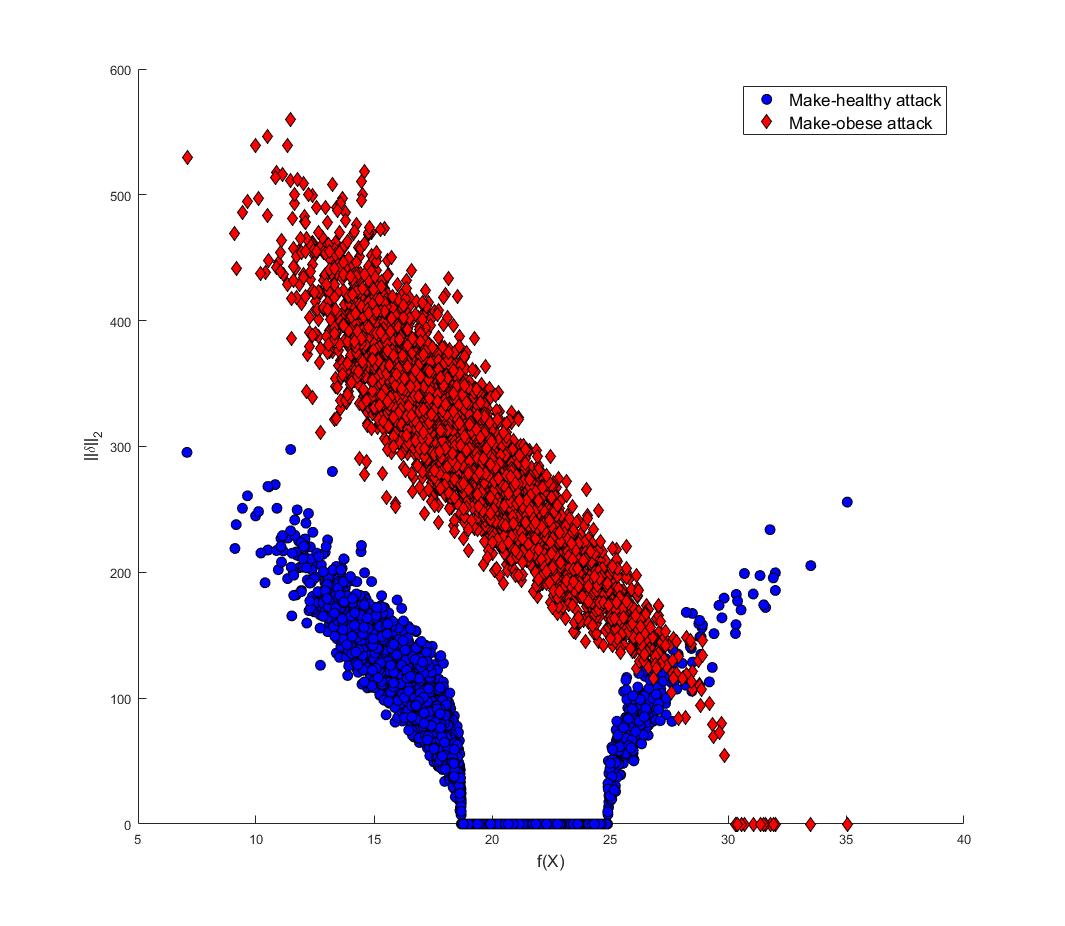}
    \caption{\scriptsize \textbf{Left:} $x$-axis: the initial BMI prediction $f(X)$, $y$-axis: the corresponding attacked BMI prediction $f(X+\delta)$ for each image in the VisualBMI data set.  We have highlighted the relevant target ranges.
    \textbf{Right:} $x$-axis: $f(X)$,  $y$-axis: the corresponding $\|\delta\|_2$ of the first successful rounded $\delta$ for each victim image in the 
    VisualBMI data set.}
    \label{fig:BMIvsAttack}
\end{figure}

\begin{figure}[htbp]
    \includegraphics[width=.49\linewidth, height = .3\linewidth,trim={0 10 10 20},clip]{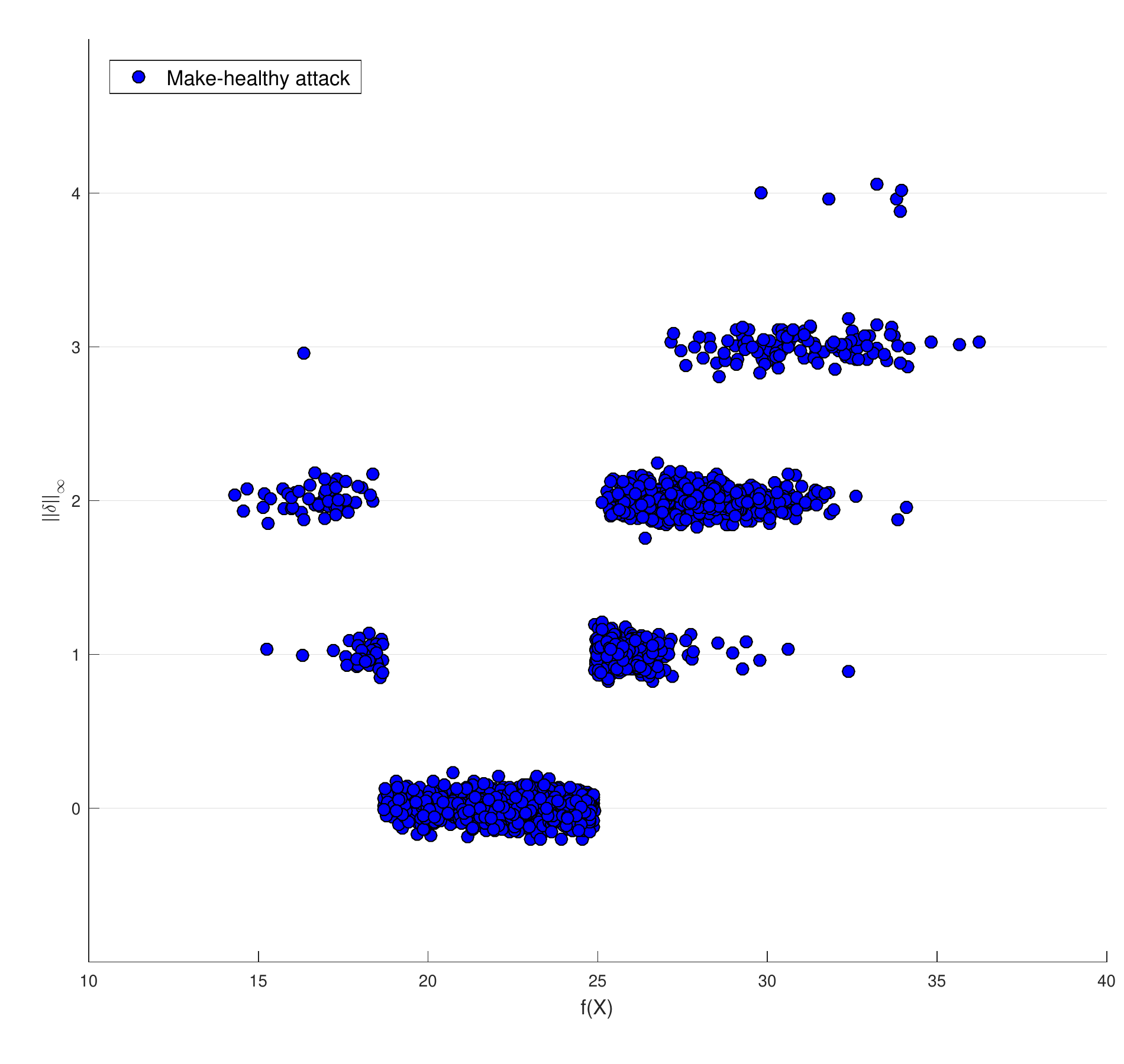}
    \includegraphics[width=.49\linewidth,height = .3\linewidth,trim={0 10 10 20},clip]{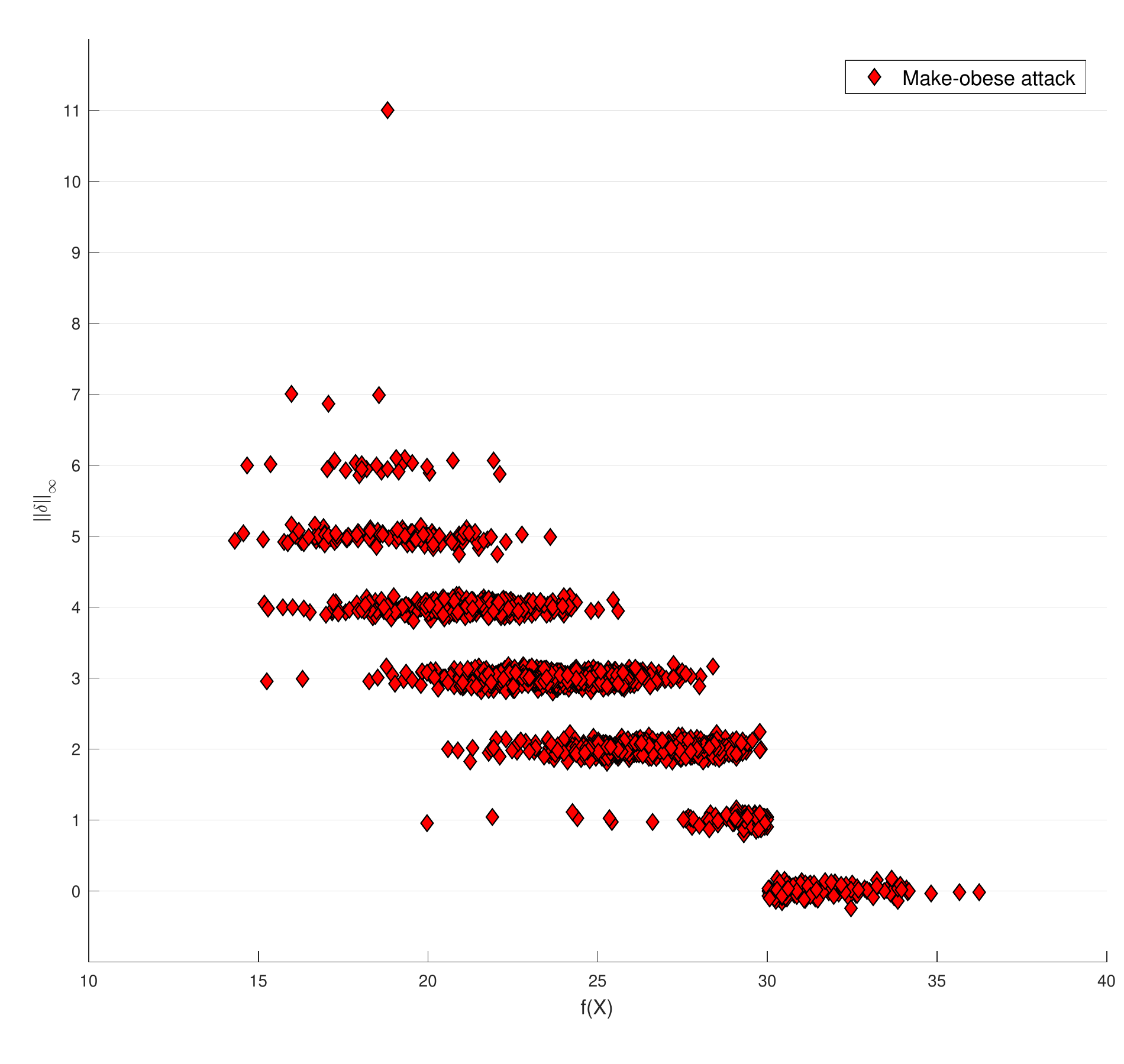}
    \caption{Attack $\|\delta\|_\infty$ on the FCBMI test set for make-healthy (Left) and make-obese (Right) attacks.
To help visualize the distribution of data we dithered the norms using iid Gaussian noise with mean 0 and variance .005}\label{fig:BMIvsNormsInf}
\end{figure}

\textbf{Quantitative results.}
We demonstrate two attacks separately: 
``make-healthy'' where the attacker forces BMI predictions into $[L,U]=[18.7,24.9]$ corresponding to normal weight, 
and ``make-obese'' with attack target range of $[L,U]=[30,40]$ corresponding to obesity.  
We use the 2045 test images from the FCBMI data set and all 4206 images in the VisualBMI data set.
Figure~\ref{fig:BMIvsAttack}(left) shows BMI before and after attack on VisualBMI.
One may expect the attack to just project the predicted BMI onto the boundary of the target range. We see almost exactly that, but there is some minor variance within the target region due to rounding of $\delta$. Infrequently, there are large outliers where the rounding shifts the prediction to the other side of the target range. One example of this phenomenon is the right-most face in Figure~\ref{fig:h}.
Figure~\ref{fig:BMIvsAttack}(right) shows $\|\delta\|_2$ under both attacks.
As expected, the further a victim's initially predicted BMI from the target region, the larger the norm of the perturbation $\delta$.
Figure~\ref{fig:BMIvsNormsInf} shows $\|\delta\|_\infty$ on the FCBMI test set.
The same trend holds.
Also note the maximum pixel value change is small, roughly 5 out of 255.
These attacks will be difficult for humans to perceive.

\section{Conclusions and future work}\label{conclusion}
We have demonstrated that na\"ive whitebox adversarial attacks can be a threat to Face-to-BMI regression.
For this reason, we urge caution when using BMI predicted from images in applications such as insurance, as they can be manipulated to make someone's rates artificially lower or higher.

The attacks in this paper requires the ability to modify any pixels. 
A more realistic attack would be physical, e.g. have the person wear make-up or accessories like glasses. 
An intermediate simulated attack could restrict the attack within face or skin pixels.  Combining these with e.g. Expectation-Over-Transformation as in \cite{athalye2017synthesizing} might allow someone to design adversarial make-up they could wear to influence the predicted BMI.

Acknowledgments:
The authors wish to thank Glenn Fung for discussions and sharing some of the data set used in this work. 
This work is supported in part by NSF 1545481, 1704117, 1836978, and the MADLab AF Center of Excellence FA9550-18-1-0166.

\bibliographystyle{alpha}
\bibliography{egbib} 
\end{document}